%% file: main.tex
\documentclass{article}
\PassOptionsToPackage{numbers,compress}{natbib}
\usepackage[preprint]{neurips_2026}
\usepackage[T1]{fontenc}
\usepackage{iftex}
\ifPDFTeX\pdfminorversion=7\fi
\ifXeTeX\AtBeginDocument{}\fi
\usepackage{amsmath,amssymb,graphicx,booktabs,array,longtable}
\usepackage{float,placeins,caption,microtype,xurl,xcolor}
\usepackage[hidelinks]{hyperref}
\hypersetup{pdftitle={CMED: Adaptive Fusion of Geometric and Learned Depth for Instrument-Scaled Endoscopy},pdfauthor={Xiaoqian Zhang and Xing Zhang}}
\title{CMED: Adaptive Fusion of Geometric and Learned Depth for Instrument-Scaled Endoscopy}
\author{%
  Xiaoqian Zhang\thanks{Work done during an internship at United Imaging Surgical.}\\
  Department of Ultrasound \& Faculty of Medical Imaging Technology\\
  Ruijin Hospital, College of Health Science and Technology\\
  Shanghai Jiao Tong University School of Medicine, Shanghai, China\\
  \texttt{xiaoqian\_shawno@sjtu.edu.cn}
  \AND
  Xing Zhang\thanks{Corresponding author.}\\
  United Imaging Surgical, Shanghai, China\\
  \texttt{xing.zhang@ui-surgical.com}
}
\begin{document}
\maketitle\raggedbottom
\input{sections/00_abstract}
\input{sections/01_introduction}
\input{sections/01_related_work}
\input{sections/02_method}
\input{sections/03_experiments}
\input{sections/04_discussion}
\FloatBarrier
\begingroup\small\input{references}\endgroup
\clearpage\appendix
\input{sections/05_appendix}
\end{document}

%% file: sections/00_abstract.tex
\begin{abstract}
Geometric reconstruction and pretrained multi-view depth networks provide complementary evidence about endoscopic surfaces, but differences in depth scale and scene-dependent errors make them difficult to combine. We present CMED, a pipeline that aligns a correspondence-based geometric depth map with a VGGT prediction through a robust multiplicative scale, predicts a single fusion weight per reference image, and recovers millimeter scale from an annotated instrument of known size. The fusion gate pairs a fully fine-tuned DINOv2 RGB encoder with a compact depth encoder and is supervised with a median-scale-aligned absolute relative error that matches the scale convention of the evaluation metric. On a group-separated development split of C3VD and SCARED, this supervision term reduces validation AbsRel from 0.10371 to 0.07884, compared with 0.10894 for fixed-weight fusion. The gain is concentrated on C3VD, while SCARED worsens slightly. A ground-truth-assisted weight search reaches 0.06422, showing that better weight selection on the same candidate maps is possible. We further apply the complete pipeline to 18 clinical image groups, producing instrument-scaled depth maps and documenting their failure modes. Evaluation on independent held-out groups and against metric clinical references is the next step before clinical accuracy can be claimed.
\end{abstract}

%% file: sections/01_introduction.tex
\section{Introduction}
Depth recovery from endoscopic images must contend with weak texture, specular appearance, restricted viewpoint changes, and moving tissue. Correspondence-based geometry ties the reconstruction to image observations and camera calibration, whereas a pretrained network supplies learned geometric priors. The two sources rarely fail in the same way, which makes their combination attractive. Combining them well requires reconciling their depth conventions and letting the mixture respond to the current scene.

CMED organizes calibrated multi-view reconstruction around a designated reference image. Here we extend the system with a parallel VGGT route and a learned, image-level fusion gate. We keep relative depth estimation, physical scale recovery, and grayscale display as separate stages, so that each can be evaluated on its own terms: a scale-aligned benchmark score measures surface shape, instrument-based calibration provides millimeters, and the display mapping is a reproducible convention rather than a source of apparent accuracy.

Our development results raise a specific question: can supervision that matches the evaluation scale convention help a learned global gate use the two routes more effectively? A ground-truth-assisted search over fusion weights reaches substantially lower validation error than the original fully fine-tuned gate, so the candidate depth maps admit better combinations than the gate was finding. Rather than enlarging the backbone, we keep the architecture and add a scale-aligned error term to the objective.

This paper makes three contributions. First, it specifies a fusion pipeline with explicit support handling and separate instrument-based scale recovery. Second, it evaluates a controlled loss change in a DINOv2-based global gate and uses source-wise results and a GT-assisted search to characterize the remaining gap. Third, it documents clinical visualization with declared calibration assumptions, a reproducible grayscale mapping, and visible failure cases. The study is exploratory: the development split contains few independent acquisition groups, the principal comparison uses one seed, and the clinical collection has no independent depth reference.

%% file: sections/01_related_work.tex
\section{Related Work}
\label{sec:related}

\subsection{Geometric and learned reconstruction in endoscopy}
Classical structure-from-motion combines camera registration, triangulation and joint refinement of structure and viewing parameters \citep{colmap,ba}. Learned correspondence estimators provide different ways to obtain the observations used by such pipelines. LoFTR uses coarse-to-fine, detector-free matching with cross-image attention \citep{loftr}. DKM estimates dense correspondences through kernel regression and refinement \citep{dkm}, while LightGlue adapts local-feature matching computation to the image pair \citep{lightglue}. These methods form the correspondence-estimation background for the RoMa-based branch of CMED.

Endoscopic depth estimation has long combined image geometry with learned priors. Liu et al.\ \citep{liuendo} use sparse supervision from multi-view reconstruction to train a dense depth network from monocular endoscopic videos. Endo-Depth-and-Motion \citep{endodepthmotion} instead obtains pseudo-RGBD observations from a self-supervised depth network, estimates camera motion through photometric tracking, and integrates registered predictions into a volumetric reconstruction. These studies show that learning and geometric constraints are complementary. CMED addresses a different step: estimating a mixing coefficient between two already computed reference-depth maps.

Self-supervised monocular learning also uses view reconstruction to supervise depth. Monodepth2 addresses occlusion and violations of camera-motion assumptions through minimum reprojection and automatic masking \citep{monodepth2}. Endoscopic illumination requires additional treatment. AF-SfMLearner models brightness variation through appearance flow and jointly estimates depth and ego-motion \citep{afsfm}. Photometric supervision in endoscopy therefore depends on acquisition conditions and image-formation assumptions.

CMED retains an explicit correspondence-based route, including camera-pose estimation, bundle adjustment, triangulation, and image-guided completion. The dense correspondences come from RoMa v2 \citep{romav2}, following the robust matching approach of RoMa \citep{roma}. A matching score measures evidence for a correspondence rather than the accuracy of the completed depth surface, so we treat the geometric reconstruction as one candidate estimate rather than as the target for the learned route.

\subsection{General-purpose visual and geometric representations}
Single-image depth methods have benefited from transferable image representations and larger training corpora. DPT reassembles Transformer features into multiscale representations for dense prediction \citep{dpt}. Depth Anything exploits large-scale unlabeled imagery \citep{depthanything}, and Depth Anything V2 combines synthetic supervision with pseudo-labeled real images \citep{depthanything2}. Marigold adapts a pretrained diffusion model to affine-invariant depth estimation \citep{marigold}. Because their training signals and scale conventions differ, aligning model outputs becomes an explicit design choice when predictions enter a reconstruction system.

Recent geometry networks reduce dependence on a conventional sequence of matching, pose recovery, and triangulation. DUSt3R \citep{dust3r} predicts pointmaps and combines multiple views through global alignment. MASt3R augments pointmap regression with local descriptors and reciprocal matching \citep{mast3r}. VGGT \citep{vggt} predicts geometric quantities from image collections; its reference-view depth head supplies CMED's second candidate map. These developments motivate comparing an explicit reconstruction with a learned geometric prior on the same reference and context images, which is the setting CMED studies.

DINOv2 \citep{dinov2} provides transferable visual representations rather than the depth prediction used in this study. We initialize the RGB branch of the fusion gate with DINOv2 ViT-S/14 and update all of its parameters, while keeping both depth-producing routes fixed. The roles are distinct: DINOv2 is the gate's image encoder, VGGT produces a candidate depth map, and RoMa v2 supplies correspondences to a separate reconstruction. Our contribution lies in how their outputs are combined.

\subsection{Adaptive fusion and scale-aware supervision}
Adaptive fusion has also been investigated specifically for endoscopic depth. CoDepth \citep{codepth} combines diffusion-based and discriminative estimators using complementary features and confidence-guided, spatially adaptive fusion. CMED builds on the same observation that depth estimators have complementary errors, but fuses a different pair of sources, explicit multi-view geometry and VGGT, and predicts one image-level scalar from RGB and aligned depth features. The scalar decision is compact, auditable, and directly comparable to a one-dimensional weight search; its trade-off is that it cannot express different local preferences within a single image. A matched comparison with CoDepth is left to future work.

Eigen et al.\ \citep{eigen} explicitly address global-scale ambiguity through a scale-invariant depth objective. Scale conventions are a further source of mismatch between a training objective and the intended depth comparison. Ranftl et al.\ \citep{midas} study learning across datasets with heterogeneous depth characteristics and objectives that accommodate scale differences. In CMED, we use a more specific convention: multiplicative median alignment, with no additive shift. Ground-truth-assisted search diagnoses how much error could be removed by changing the global coefficient on the available candidates. It motivates the aligned AbsRel term used in training; the searched weights themselves are neither deployment inputs nor regression targets. The contribution examined here is this controlled supervision change and its source-wise behavior.

\subsection{Depth references and metric interpretation}
C3VD provides paired depth for colonoscopy models through 2D--3D registration \citep{c3vd}; SCARED provides endoscopic reconstruction data acquired from porcine cadavers \citep{scared}. They support quantitative development where reference depth is available; neither represents our human clinical collection. Because comparisons across papers depend on acquisition groups, valid-pixel support, scale alignment, and aggregation, we report our local subset and validation protocol explicitly.

Finally, relative-shape accuracy and physical scale are separate questions. Our public-data metric removes a global scale using ground truth; deployment instead estimates scale from an instrument span and calibrated camera rays. The latter depends on correct annotation correspondence and the stated geometric assumptions. The clinical examples therefore illustrate supported reconstruction and conditional metric visualization; independent millimeter-accuracy validation is a separate step.

%% file: sections/02_method.tex
\section{Method}
\label{sec:method}
\begin{figure}[!htbp]
\centering\includegraphics[width=\linewidth]{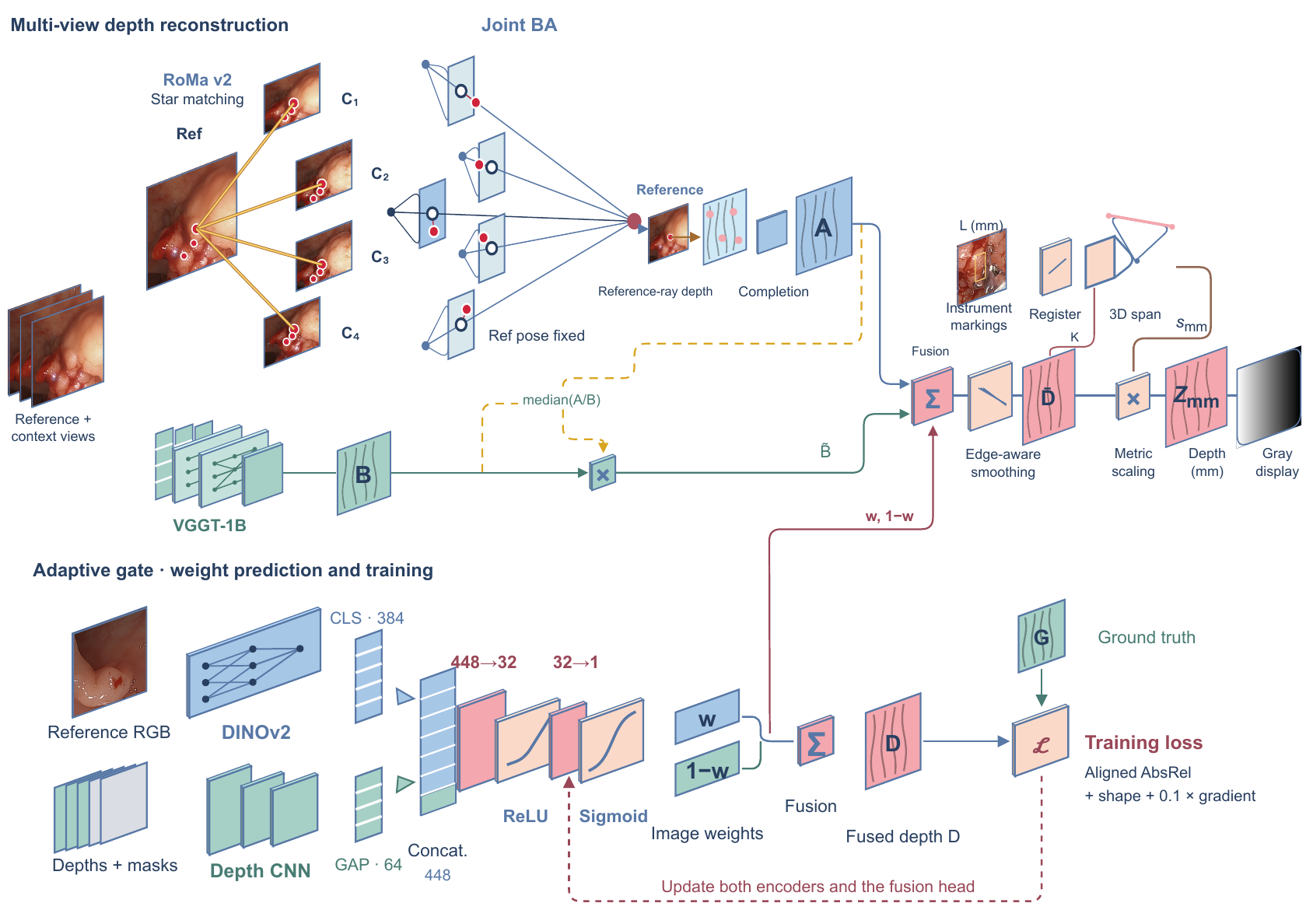}
\caption{Multi-view endoscopic depth reconstruction and image-level fusion. Top: RoMa v2 matches each context to the reference. Gold spokes illustrate one track; red dots show observed pixels and open circles show reprojections in joint bundle adjustment with a fixed reference pose. Reference-ray depth estimation and completion yield $A$. VGGT predicts $B$ from the same views. On common support, the median ratio aligns $B$ to $A$, giving $\widetilde B$. The gate supplies $w$ for $A$ and $1-w$ for $\widetilde B$. Edge-aware smoothing yields $\bar D$, and a registered instrument span with camera intrinsics sets the millimeter scale. Gray values show offsets from the nearest supported depth. Bottom: DINOv2 encodes reference RGB and a CNN encodes the depths and support masks. The head predicts one weight pair per reference image. The $\Sigma$ symbols denote the same fusion operation. Training compares its output $D$ with ground truth $G$ before smoothing. The dashed path indicates updates to the gate encoders and head. RGB crops, sampled matches and instrument markings come from the clinical acquisition. Depth and feature planes are schematic.}
\label{fig:method}
\end{figure}

\subsection{Inputs and coordinate contract}
The inputs are an annotated instrument observation with a known physical span, a designated reference image, and context views of the same local scene. Known camera intrinsics and distortion parameters define a single undistortion into a common pixel grid. The reference image remains frame 0 even if another view initializes geometry. All depths below denote positive reference-camera optical-axis coordinates $Z$, not Euclidean distance to the camera center. For an undistorted pixel $p=(u,v)$, its camera ray is $r_p=K^{-1}[u,v,1]^\top$ and its point is $X_p=Z(p)r_p$.

\subsection{The existing geometric reconstruction route}
\label{sec:geometry}
The upper branch of Fig.~\ref{fig:method} expands the existing route into stages with distinct outputs. Geometry is computed before gate learning; the gate does not modify the matches, camera poses, or completed candidate depths.

\paragraph{Dense matching and observation filtering.}
We build a star-shaped matching graph centered on the fixed reference view $I_0$. For each input context view $I_i$, RoMa v2.0.1 in its precise setting processes the pair $(I_0,I_i)$. The edge set is $\{(0,i): i=1,\ldots,N\}$ for $N$ context views. Observations at the same reference pixel form one multi-view track across the contexts where the correspondence passes filtering. A retained observation must be finite, inside the image domain, and satisfy the configured overlap, positional-uncertainty, and forward--backward consistency checks. Reference and target masks also constrain correspondences. The resulting tracks carry pixel coordinates and overlap-based weights. These checks belong to the reconstruction stage; the later fusion stage introduces no additional confidence or disagreement rejection.

\paragraph{Camera initialization and bundle adjustment.}
Candidate reference--context pairs undergo essential-matrix RANSAC, pose recovery, and sparse triangulation. The initialization pair is selected using track visibility in the remaining views. PnP-RANSAC followed by refinement initializes those views. A Huber-robust, weighted bundle adjustment then jointly refines their poses and sparse 3D points against image reprojection residuals. We write $X_i=R_iX_b+t_i$, so $T_{i\leftarrow b}$ maps the reference-camera frame into context camera $i$. The reference pose is fixed to identity and the anchor translation has unit norm. This fixes a reconstruction gauge, not a metric baseline. Insufficient parallax or too few PnP correspondences can prevent reconstruction; in that case the clinical runner records the failing context view and may remove it before retrying when at least three views remain. The reference image is retained, and the resulting retained-view indices are also used by VGGT.

\paragraph{Confidence-weighted reference-ray depth.}
With poses fixed after BA, dense depth is recomputed along each reference ray $r_p$. Let $q_{ip}=K^{-1}[u_{ip},v_{ip},1]^\top$ be the matched context ray, and define
\begin{align}
 a_{ip}&=q_{ip,xy}(R_i r_p)_z-(R_i r_p)_{xy},\\
 b_{ip}&=t_{i,xy}-q_{ip,xy}t_{i,z}.
\end{align}
The implemented weighted algebraic least-squares solution is
\begin{equation}
 Z_p^*=\arg\min_Z\sum_i c_{ip}\|a_{ip}Z-b_{ip}\|_2^2
 =\frac{\sum_i c_{ip}a_{ip}^{\top}b_{ip}}
 {\sum_i c_{ip}\|a_{ip}\|_2^2},
\label{eq:multiray}
\end{equation}
where $c_{ip}$ is the retained overlap weight. The denominator must exceed the numerical tolerance; only finite positive reference-camera depths become seeds. No additional dense RANSAC is applied at this stage. Reprojection agreement and seed density serve as geometric diagnostics.

\paragraph{Anchored inverse-depth completion.}
The seed field is converted to $\rho_p=1/Z_p^*$. Completion solves an image-guided weighted Laplacian/Poisson system on a four-neighbor pixel graph, with grayscale differences controlling edge weights. Seed inverse depths are fixed boundary values. Unknowns are restricted to the permitted image domain, within the configured seed-distance radius, and in fill components connected to seeds. The solution is converted back to positive depth to produce $A$ and support $M_A$. Completion therefore interpolates within supported neighborhoods and does not fill every unsupported cavity. Weak texture, specular appearance, or erroneous camera poses propagate into the completed surface.

\subsection{The parallel VGGT route}
The second route passes the preprocessed views directly to the original VGGT-1B aggregator. Its internal DINOv2 ViT-L/14 image encoder produces patch tokens for alternating frame and global attention. The VGGT depth head decodes the aggregated tokens. The separate DINOv2 ViT-S/14 in the fusion gate is described below. The same retained, undistorted image sequence is supplied, and the reference prediction is mapped back to the geometric route's pixel grid. Public caches use $640\times512$ images, with VGGT preprocessing to $518\times420$ and bilinear mapping back; the adapter resizes rather than crops. Clinical exports retain their recorded per-image preprocessing transforms. This branch outputs $B$ and its finite-positive support $M_B$; it does not consume the geometric BA poses or instrument annotations. Both routes use the same reference-camera depth convention but need not share a numerical scale. Their outputs are cached and remain fixed throughout the gate experiments.

\subsection{GT-free alignment and adaptive fusion}
Let $A$ and $B$ denote the geometric and VGGT depths, with supports $M_A$ and $M_B$. Support includes the image domain and finite positive values. The common support $\Omega=M_A\cap M_B$ must be nonempty. We put VGGT in the geometric scale using
\begin{equation}
 s_B=\operatorname{median}_{p\in\Omega}\frac{A(p)}{B(p)},\qquad \widetilde B=s_B B.
\end{equation}
This is a single multiplicative transformation, with no additive shift and no access to GT.

The gate in the bottom row of Fig.~\ref{fig:method} predicts $w_\theta\in(0,1)$ from RGB and depth features. With $m_A=\operatorname{median}_{M_A}A$, the depth channels are $\log(A/m_A)$, $\log(\widetilde B/m_A)$, their absolute difference on $\Omega$, and the two support masks. Unsupported channel entries are zero. A shared normalizer preserves the relation between the two depths. Inputs are letterboxed to $256\times320$; RGB uses area resampling and depth/mask channels use nearest sampling.

A pretrained DINOv2 ViT-S/14 encoder supplies a 384-dimensional RGB CLS representation. ImageNet normalization and replicate padding produce a $266\times322$ patch-compatible input. A three-layer depth CNN ($5\!\to\!16\!\to\!32\!\to\!64$, $3\times3$, stride 2, ReLU) followed by global average pooling supplies 64 features. Concatenation gives 448 features, followed by a $448\!\to\!32\!\to\!1$ ReLU/sigmoid head. The full model has 21,736,577 trainable parameters. All DINOv2 parameters are updated in the selected variant; the two depth-producing models remain fixed. The final head initializes $w_\theta=0.7$.

The unsmoothed fusion is
\begin{equation}
 D(p)=\begin{cases}
 w_\theta A(p)+(1-w_\theta)\widetilde B(p),&p\in M_A\cap M_B,\\
 A(p),&p\in M_A\setminus M_B,\\
 \widetilde B(p),&p\in M_B\setminus M_A.
 \end{cases}
\label{eq:fusion}
\end{equation}
Pixels outside the union remain unavailable. The scalar gate adapts across images; within an image, the same weight applies everywhere.

For deployment, let $\bar D$ denote the postprocessed depth. Normalized bilateral averaging is restricted to single-route regions and a two-pixel band around source transitions. Within a $7\times7$ neighborhood, spatial distance, RGB difference, and log-depth difference determine weights, with scales 2 pixels, 0.1 in RGB $[0,1]$, and 0.1 in log depth. Only available neighbors contribute, which limits propagation across appearance and depth edges. All public gate-selection results in this paper are measured \emph{before} this postprocessing, so its effect is never attributed to the learned gate.

\subsection{Supervision aligned with the evaluation scale}
GT depth $G$ is available only for supervised public-data training and offline evaluation. The loss support is $\Omega_G=M_A\cap M_B\cap M_G$. With $e_p=\log D(p)-\log G(p)$, the baseline loss is the mean centered squared log residual plus $0.1$ times the mean absolute horizontal and vertical differences of $e$ over supported neighbor pairs. Denoting these terms by $\mathcal L_{\rm shape}$ and $\mathcal L_{\rm grad}$, the updated loss is
\begin{align}
 s_G&=\operatorname{median}_{p\in\Omega_G}\frac{G(p)}{D(p)},\\
 \mathcal L_{\rm aligned}&=\frac{1}{|\Omega_G|}\sum_{p\in\Omega_G}\frac{|s_GD(p)-G(p)|}{G(p)},\\
 \mathcal L&=\mathcal L_{\rm shape}+0.1\mathcal L_{\rm grad}+1.0\mathcal L_{\rm aligned}.
\label{eq:loss}
\end{align}
For an even number of pixels, the two central sorted ratios are averaged. Gradients propagate through the selected median values; the scale is not detached. The added term uses the same pixel error and ratio-based median convention as evaluation. Training averages images on a nearest-sampled $256\times320$ grid, whereas evaluation uses full-resolution predictions and equal sequence weighting, so the two objectives match in scale convention but not in sampling or aggregation.

The GT-assisted diagnostic evaluates $w\in\{0,0.01,\ldots,1\}$ independently for each image under the same metric. Its minimum is a finite-grid reference rather than a continuous lower bound or a deployable method. Oracle weights are not network inputs or regression targets, and validation GT is never used in gradient updates.

\subsection{Instrument-based millimeter scale and display}
After relative fusion and boundary smoothing, the depth is $\bar D$. An annotated segment is mapped into the undistorted reference grid. If its endpoints $p_1,p_2$ both have supported depth, a known physical span $L_{\rm mm}$ sets
\begin{equation}
 s_{\rm mm}=\frac{L_{\rm mm}}{\|\bar D(p_2)r_{p_2}-\bar D(p_1)r_{p_1}\|_2},\qquad Z_{\rm mm}(p)=s_{\rm mm}\bar D(p).
\end{equation}
The span must correspond to the same physical endpoints as the annotation; incorrect endpoint association, registration, or camera calibration scales the entire map.

If an endpoint lacks depth, the fallback samples 101 positions along the \emph{original marked segment} and requires at least 10 supported samples. It estimates a median relative anchor and assumes equal camera-axis depth along the segment. With $\delta r=\|r_{p_2,xy}-r_{p_1,xy}\|_2$, the estimated metric anchor is $L_{\rm mm}/\delta r$; dividing it by the relative anchor gives $s_{\rm mm}$. Instrument tilt or unrepresentative visible support violates this approximation. Both modes yield conditional scale estimates.

Floating-point depth is retained in millimeters. The requested grayscale display uses each map's nearest supported depth $Z_{\min}$:
\begin{equation}
 g(p)=\operatorname{round}\!\left[\operatorname{clip}\!\left(255-10[Z_{\rm mm}(p)-Z_{\min}],16,255\right)\right].
\end{equation}
Unavailable pixels are 0. Ten gray-value steps therefore correspond to one millimeter. Values beyond 23.9 mm from the nearest point saturate at 16 in the display only. Equal gray values across images represent equal offsets from their respective nearest depths, \emph{not} equal absolute depths; the underlying floating-point estimates are unchanged.

%% file: sections/03_experiments.tex
\section{Experimental setup and development results}
\label{sec:experiments}
\subsection{Data, splits, and evaluation}
We use a local, group-separated subset of C3VD and SCARED. The frozen manifest contains 150 four-view examples divided 90/30/30 into training, validation, and a previously evaluated test partition (3:1:1). Splitting is by physical geometry or acquisition/subject group rather than by random frame. C3VD contributes 100 examples from three geometries: sigmoid for training, cecum for validation, and transverse colon for test. SCARED contributes 50 context configurations but only five distinct labeled reference keyframes. The latter are grouped by acquisition identifiers 6/7/3 for training, 2 for validation, and 1 for test. Table~\ref{tab:split} distinguishes example counts from independent grouping units.

\begin{table}[!htbp]
\centering\caption{Local public-data subset. Entries are examples (geometry or acquisition groups). Repeated contexts are not independent labeled subjects.}\label{tab:split}
\begin{tabular}{lrrr}\toprule
Source & Train & Validation & Prior test partition\\\midrule
C3VD & 60 (1) & 20 (1) & 20 (1)\\
SCARED & 30 (3) & 10 (1) & 10 (1)\\\midrule
Total & 90 (4) & 30 (2) & 30 (2)\\\bottomrule
\end{tabular}
\end{table}

The earlier small-CNN experiments examined the test partition. All subsequent DINOv2 development, the added loss, and model selection used training and validation only, so we report the new model as a \emph{development validation result} and reserve held-out testing for a sealed collection. Twelve contexts with failed pose initialization were deterministically replaced with earlier RGB contexts before initial gate training; reference RGB, GT, and split assignment were unchanged, and no examples were dropped.

C3VD depth values at its 100 mm encoding cap are excluded from GT support; SCARED uses its released XYZ-derived camera $Z$ and source-specific calibration without applying that cap. All methods are evaluated on exactly the common valid depth support with GT. For every image and every candidate, an evaluation-only scale $\operatorname{median}(G/D)$ precedes AbsRel. Errors are averaged within each sequence and then equally across sequences. With one validation sequence per source, the reported macro average is the average of the two source values. The score assesses relative depth shape on common support; the instrument-based scale estimator is assessed separately.

\paragraph{Optimization and controls.}
All principal gate comparisons use seed 42, 200 epochs, AdamW, batch size 16, weight decay $10^{-4}$, gradient clipping at 5, and cosine learning-rate decay. The fusion head learning rate is $10^{-3}$ and the trainable DINOv2 learning rate is $10^{-5}$. Training uses FP32 on a local Apple MPS device. A checkpoint is selected by the lowest validation sequence-macro AbsRel over all 200 epochs. The added-loss comparison keeps the full architecture, route caches, initialization, data split, and optimization schedule fixed. An inference checkpoint reloaded independently on CPU reproduces the training-device score (0.07883516 versus 0.07883515).

\subsection{Task-aligned supervision improves the aggregate, with source dependence}
\input{tables/validation}
\begin{figure}[!htbp]\centering
\includegraphics[width=\linewidth]{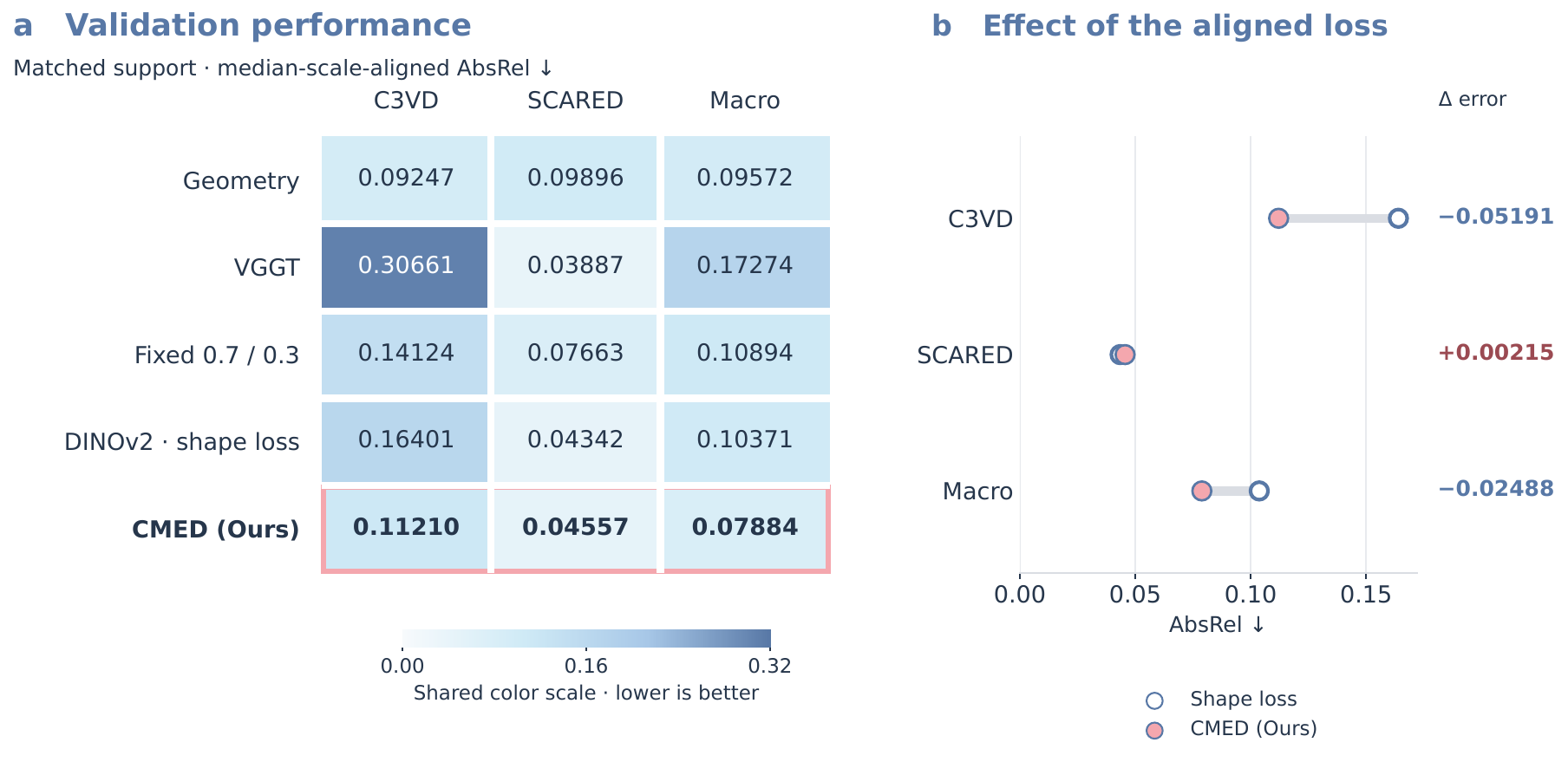}
\caption{Validation AbsRel on matched common support; lower is better. CMED (Ours) denotes the fully fine-tuned gate with aligned AbsRel supervision. These scores assess its relative-depth fusion output before smoothing and instrument calibration. \textbf{(a)} The annotated matrix uses one color scale across sources. \textbf{(b)} Paired markers isolate the change from shape-only to aligned-loss supervision; $\Delta$ is updated minus baseline error. The updated gate improves the aggregate relative to its shape-loss counterpart, while SCARED worsens slightly. Geometry alone remains better on C3VD, and VGGT alone remains better on SCARED. All values are development validation comparisons.}
\label{fig:validation}\end{figure}

Adding aligned AbsRel supervision reduces macro error from 0.10371 to 0.07884, a 23.99\% relative reduction. The selected epoch changes from 12 to 22. Fixed 0.7/0.3 fusion scores 0.10894, and geometry alone scores 0.09572. The updated gate therefore gives the lowest aggregate error among the deployable variants on this validation split.

The source breakdown refines this picture. C3VD improves from 0.16401 to 0.11210, while SCARED moves from 0.04342 to 0.04557. The fused result remains above geometry alone on C3VD (0.09247) and VGGT alone on SCARED (0.03887), so the gate finds a better compromise across the mixture rather than dominating either source on its own data. With two validation grouping units and one seed, we report these values descriptively without confidence intervals.

\begin{figure}[!htbp]\centering
\includegraphics[width=\linewidth]{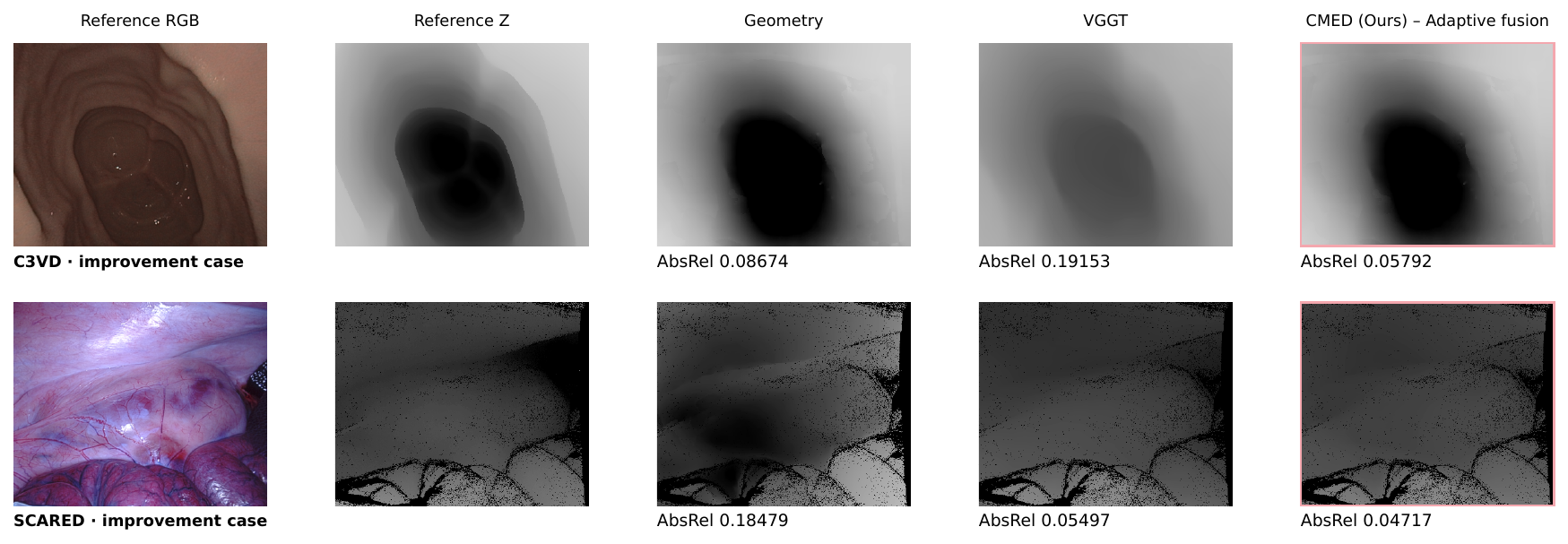}
\caption{Illustrative improvement cases from public validation. C3VD (\texttt{cecum\_t1\_a\_0076}) improves over Geometry; SCARED (\texttt{2\_key2\_s4}) improves over VGGT. Each is the single case meeting its improvement criterion among 20 C3VD and 10 SCARED validation images, respectively. CMED (Ours)--Adaptive fusion denotes the unsmoothed relative-depth fusion output, before instrument calibration. Predictions are separately GT-median aligned and evaluated on identical common support; each row shares a display window ending at the 98th GT percentile. Alignment is evaluation-only.}
\label{fig:public}\end{figure}

\subsection{Gate ablations and GT-assisted headroom}
\input{tables/ablation}
Table~\ref{tab:ablation} separates backbone training from the added objective. A frozen pretrained encoder does not improve the seed-42 small-CNN result. Updating the last two Transformer blocks and final normalization improves upon the frozen configuration; full updating improves further. The largest reduction in this sequence follows the controlled objective change, not the increase in trainable capacity. Because the backbone variants change both representation and capacity, the table does not isolate a pure size effect.

\begin{figure}[!htbp]\centering
\includegraphics[width=\linewidth]{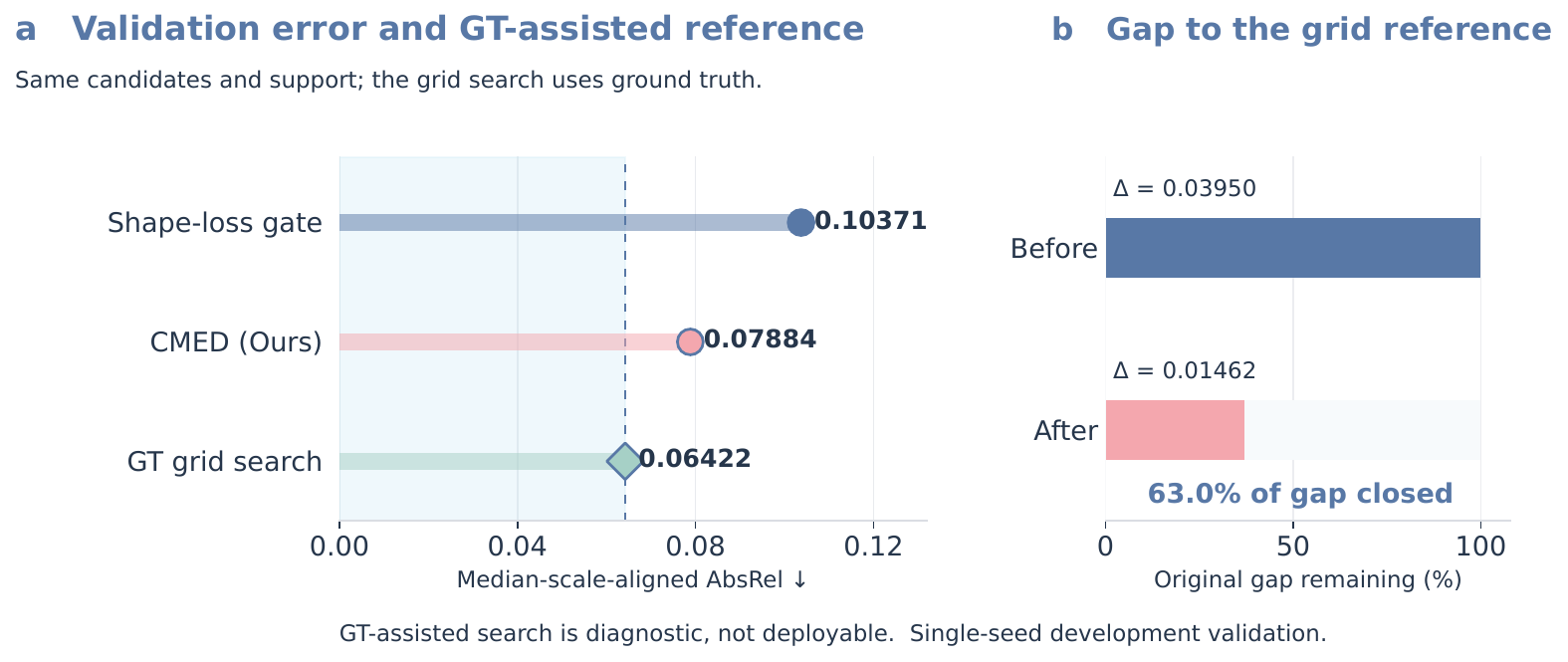}
\caption{A GT-assisted finite-grid diagnostic exposes remaining fusion-selection headroom. \textbf{(a)} Errors on the same validation cases and metric; the diamond denotes a GT-assisted diagnostic, not a deployable competitor. \textbf{(b)} The gap to this reference falls from 0.03950 to 0.01462: 63.0\% of the original gap is closed, leaving 37.0\%. Values are from the single-seed development run.}
\label{fig:oracle}\end{figure}

The 101-point weight search reaches 0.06422 on validation, compared with 0.07884 for the updated gate. Its gap decreases from 0.03950 for the shape-loss gate to 0.01462, approximately a 63.0\% reduction. At the updated checkpoint, training error is 0.06685 versus a grid reference of 0.06319 (gap 0.00366). The larger validation gap points to remaining weight-selection difficulty on unseen groups. A continuous predicted weight can occasionally beat the discrete grid; such negative per-image gaps are retained rather than clipped.

\subsection{Clinical validation of instrument-calibrated depth}
The private collection consists of 18 clinical image groups from Jiangsu Province Hospital (The First Affiliated Hospital of Nanjing Medical University), Nanjing, China. The supplied instrument annotations and physical sizes are used only for conditional metric scale recovery. We apply the complete CMED (Ours) pipeline, including adaptive relative-depth fusion, restricted smoothing, and instrument-based scale recovery, to all 18 groups. The resulting floating-point depth maps are expressed in millimeters. Because no per-pixel clinical depth GT exists for this collection, this evaluation assesses pipeline applicability and qualitative reconstruction behavior. Group P310 was excluded before export.

\begin{figure}[!htbp]\centering
\includegraphics[width=.86\linewidth]{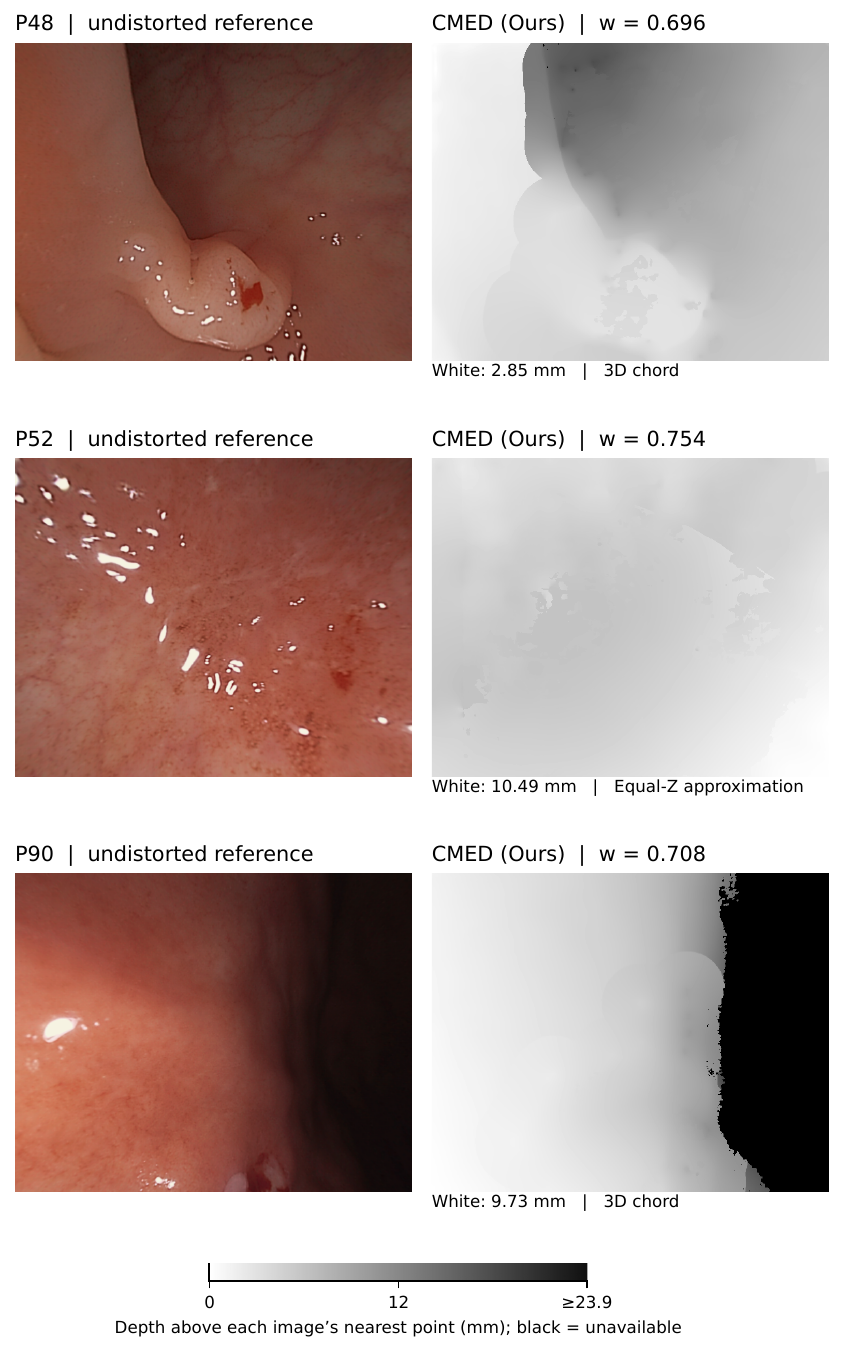}
\caption{Instrument-calibrated millimeter-depth estimates from CMED (Ours) on three visually readable examples from the 18-group Jiangsu Province Hospital collection. P48 shows a local surface/recess with chord calibration; P52 includes a low-texture, low-contrast result and equal-$Z$ fallback; P90 shows the supported wall adjacent to an unavailable lumen region. The examples were chosen for display readability and coverage of these situations. Each row starts at its own nearest depth; equal gray values across rows do not imply equal absolute $Z$.}
\label{fig:clinical}\end{figure}

The gate's geometric-route weight ranges from 0.146 to 0.800 across the 18 groups. Thirteen use endpoint-based 3D chord scaling and five use the equal-$Z$ fallback (P7, P12, P52, P62, P202). P307 retains provisional annotation registration. The examples in Fig.~\ref{fig:clinical} were selected for readable structure after examining all maps; the complete collection is in Appendix~\ref{app:clinical}, where circular completion artifacts, banding, unavailable regions, and grayscale saturation are also documented.

%% file: tables/validation.tex
\begin{table}[!htbp]
\centering\caption{Development validation AbsRel $\downarrow$, on matched common support before smoothing. Seed 42 for learned gates; selected using validation. The GT-assisted row is a non-deployable diagnostic and is excluded from model ranking.}\label{tab:validation}
\begin{tabular}{lrrr}\toprule
Method & C3VD & SCARED & Macro\\\midrule
Geometry & 0.09247 & 0.09896 & 0.09572 \\
VGGT & 0.30661 & 0.03887 & 0.17274 \\
Fixed 0.7 / 0.3 & 0.14124 & 0.07663 & 0.10894 \\
DINOv2 full, shape loss & 0.16401 & 0.04342 & 0.10371 \\
CMED (Ours) & 0.11210 & 0.04557 & 0.07884 \\
\midrule
GT-assisted grid search & 0.09046 & 0.03797 & 0.06422 \\
\bottomrule
\end{tabular}
\end{table}

%% file: tables/ablation.tex
\begin{table}[!htbp]\centering
\caption{Seed-42 gate ablations, each trained for 200 epochs and selected over all epochs. Only the last row adds aligned AbsRel to the shape/gradient objective. Frozen refers to the RGB encoder; the fusion head remains trainable.}\label{tab:ablation}
\begin{tabular}{lrrr}\toprule
Gate & Trainable parameters & Epoch & Val. AbsRel\\\midrule
Small RGB-depth CNN & 26,417 & 1 & 0.10917 \\
DINOv2 frozen + head & 38,273 & 6 & 0.11300 \\
DINOv2 last two blocks + head & 3,589,505 & 11 & 0.10836 \\
DINOv2 full, shape loss & 21,736,577 & 12 & 0.10371 \\
CMED (Ours) & 21,736,577 & 22 & 0.07884 \\
\bottomrule
\end{tabular}
\end{table}

%% file: sections/04_discussion.tex
\section{Discussion and limitations}
The central development result is that a gate trained only with centered log-shape and gradient terms leaves substantial headroom under the median-scale-aligned AbsRel metric, and that adding the metric's scale convention to the supervision recovers much of it without changing the candidate depth maps or the model capacity. The GT-assisted search is a useful diagnostic for this headroom. Whether RGB and the two predicted maps carry enough information to identify the optimal weight at inference remains open.

A global weight is simple to interpret and audit, but it cannot resolve spatially varying route quality within one image. Missing support is a separate issue: the gate is supervised and scored on the common support, while deployment also fills single-source regions and applies restricted smoothing. The validation improvement therefore speaks to the fused shape on common support, not to those regions, the smoothing policy, or downstream millimeter calibration. Future comparisons should evaluate coverage and error together and test spatial adaptation with explicit controls.

The data impose a stronger limitation than the raw example count suggests. Validation includes one C3VD geometry and one SCARED acquisition group, with repeated reference content in SCARED, and this split has guided several model and objective decisions. The updated gate has not yet been evaluated on a sealed, independent group-level test collection or across multiple seeds; the earlier test partition was used by the small-CNN baseline and is not treated as untouched. We therefore present the improvement as a development result that motivates such an evaluation, and we make no cross-hospital or population-level generalization claim.

Physical units add further assumptions. The calibration segment must identify the correct physical span in the reference frame, the equal-$Z$ fallback is sensitive to instrument orientation, and image-to-image annotation registration can be inaccurate. Camera calibration transfer, tissue motion, and completion errors affect both shape and scale. The instrument-derived factor therefore yields a conditional metric estimate pending independent validation. Clinical grayscale appearance also depends on the per-image origin and saturation, so numerical comparisons should use the floating-point depth files.

\section{Conclusion}
CMED combines geometric and pretrained multi-view depth through a fully fine-tuned DINOv2 global gate and recovers millimeter scale from instrument annotations. Task-aligned supervision lowers the aggregate development validation error, and source-wise results together with a GT-assisted diagnostic locate the remaining gap. Public-data shape evaluation, instrument-based scaling, and clinical visualization are reported as separate stages. Independent grouped testing, repeated seeds, and external metric references are the natural next steps toward clinically validated depth accuracy.

%% file: references.tex

%% file: sections/05_appendix.tex
\section{Clinical calibration and display record}
\input{tables/clinical}
\FloatBarrier\clearpage
\section{Complete 18-group qualitative record}
\label{app:clinical}
\begin{figure}[H]\centering
\includegraphics[width=\linewidth]{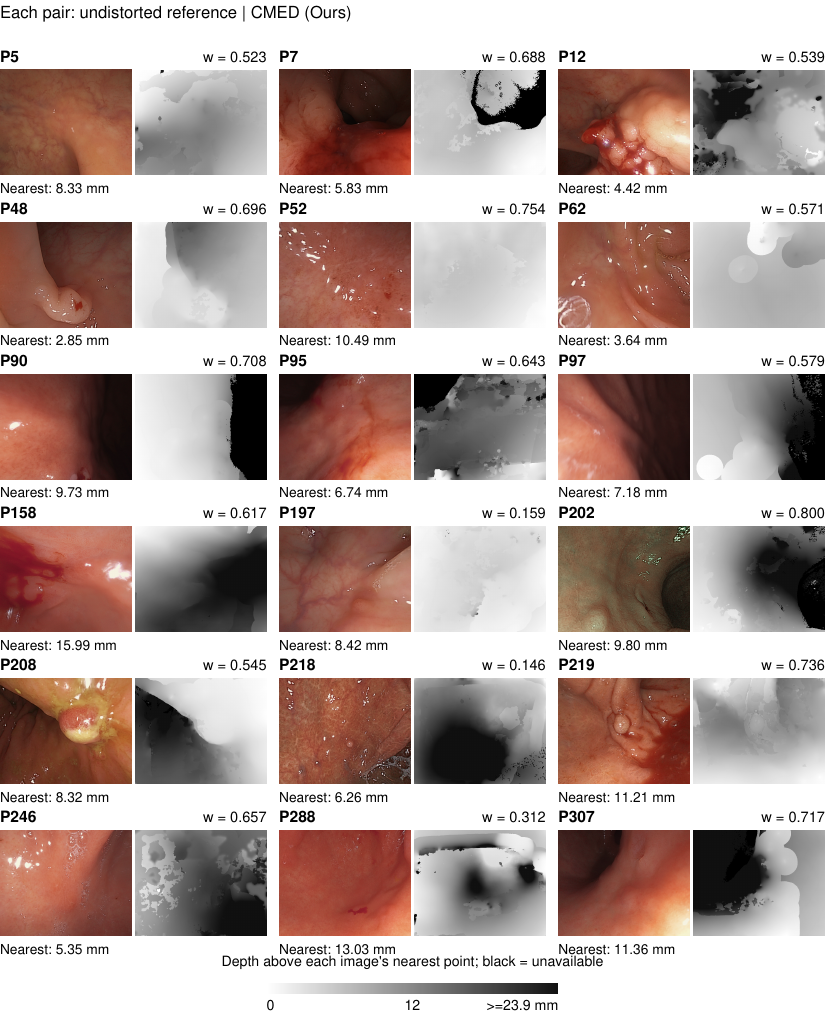}
\caption{Complete 18-group clinical record in one composite figure, read left to right and top to bottom. Each pair shows the undistorted reference image (left) and CMED depth (right). White marks each image's nearest estimated depth; intensity decreases by ten gray values per millimeter to a valid minimum of 16. Black denotes unavailable depth. All source image pixels and all included cases are preserved; only layout and display size change.}
\label{fig:clinical_all18}
\end{figure}

Figure~\ref{fig:clinical_all18} shows all 18 clinical image groups from Jiangsu Province Hospital in identifier order. The calibration values and modes are listed in Table~\ref{tab:clinical}. These are conditional instrument-scaled estimates without independent clinical ground truth.

%% file: tables/clinical.tex
\begin{table}[H]\centering\small
\caption{All 18 clinical groups. Coverage is the available fraction of the image grid. Saturation is the fraction of available pixels exceeding the 23.9 mm display-offset span. Chord and equal-$Z$ are conditional calibration modes.}\label{tab:clinical}
\begin{tabular}{lrrrrl}\toprule
Group & $w$ & Nearest (mm) & Coverage (\%) & Saturated (\%) & Scale mode\\\midrule
P5 & 0.523 & 8.33 & 100.0 & 0.0 & Chord \\
P7 & 0.688 & 5.83 & 85.6 & 0.0 & Equal-$Z$ \\
P12 & 0.539 & 4.42 & 99.8 & 8.4 & Equal-$Z$ \\
P48 & 0.696 & 2.85 & 100.0 & 0.0 & Chord \\
P52 & 0.754 & 10.49 & 100.0 & 0.0 & Equal-$Z$ \\
P62 & 0.571 & 3.64 & 100.0 & 0.0 & Equal-$Z$ \\
P90 & 0.708 & 9.73 & 74.2 & 0.0 & Chord \\
P95 & 0.643 & 6.74 & 91.6 & 1.1 & Chord \\
P97 & 0.579 & 7.18 & 83.4 & 5.1 & Chord \\
P158 & 0.617 & 15.99 & 100.0 & 12.6 & Chord \\
P197 & 0.159 & 8.42 & 100.0 & 0.0 & Chord \\
P202 & 0.800 & 9.80 & 90.3 & 11.7 & Equal-$Z$ \\
P208 & 0.545 & 8.32 & 99.3 & 4.6 & Chord \\
P218 & 0.146 & 6.26 & 99.5 & 22.5 & Chord \\
P219 & 0.736 & 11.21 & 100.0 & 0.0 & Chord \\
P246 & 0.657 & 5.35 & 100.0 & 10.7 & Chord \\
P288 & 0.312 & 13.03 & 100.0 & 2.6 & Chord \\
P307 & 0.717 & 11.36 & 95.7 & 29.8 & Chord (provisional) \\
\bottomrule
\end{tabular}
\end{table}